\documentclass{Interspeech2024}
\usepackage{subcaption}

\interspeechcameraready

\title{Uncertainty-Aware Mean Opinion Score Prediction}

\name[affiliation={1}]{Hui}{Wang}
\name[affiliation={1}]{Shiwan}{Zhao}
\name[affiliation={1}]{Jiaming}{Zhou}
\name[affiliation={2}]{Xiguang}{Zheng}
\name[affiliation={1}]{Haoqin}{Sun}
\name[affiliation={1}]{Xuechen}{Wang}
\name[affiliation={1,*}]{\\Yong}{Qin}

\address{
  $^1$Nankai University, China\\
  $^2$University of Wollongong, Australia }
\email{qinyong@nankai.edu.cn}


\begin{document}

\maketitle
\renewcommand{\thefootnote}{}
\footnotetext{* Corresponding author.}
\begin{abstract}
    
    Mean Opinion Score (MOS) prediction has made significant progress in specific domains. However, the unstable performance of MOS prediction models across diverse samples presents ongoing challenges in the practical application of these systems. In this paper, we point out that the absence of uncertainty modeling is a significant limitation hindering MOS prediction systems from applying to the real and open world. We analyze the sources of uncertainty in the MOS prediction task and propose to establish an uncertainty-aware MOS prediction system that models aleatory uncertainty and epistemic uncertainty by heteroscedastic regression and Monte Carlo dropout separately. The experimental results show that the system captures uncertainty well and is capable of performing selective prediction and out-of-domain detection. Such capabilities significantly enhance the practical utility of MOS systems in diverse real and open-world environments.

\end{abstract}

\section{Introduction}

With the advancement of deep learning, particularly self-supervised learning (SSL), MOS prediction systems have made notable progress in specific domains \cite{lo19_interspeech, 9413877, 9747222, 9746395, 10219692}. For instance, Saeki et al. \cite{saeki22c_interspeech} propose an ensemble learning method combining strong SSL learners with weak learners, achieving the highest scores across several metrics in the VoiceMOS Challenge 2022 \cite{huang22f_interspeech}.  Wang et al. \cite{wang23r_interspeech} introduce a retrieval method to enhance the MOS prediction system and design a fusion network to optimize the retrieval scope and fusion weights.

While MOS prediction systems have demonstrated commendable overall performance, their deployment in real-world, open environments poses significant challenges. The inconsistency in performance across various samples within the dataset, coupled with a pronounced decline in effectiveness when encountering zero-shot out-of-domain (OOD) samples \cite{tseng22b_interspeech}, underscores the limitations of current systems. Active research is underway to address these issues. Qi et al. \cite{10389788} innovatively integrate supervised LE-SSL-MOS with unsupervised methods, marking a significant advancement in evaluating the OOD French synthesized speech, and the noisy and enhanced speech. Shen et al. \cite{10389681} develop a speech quality assessment transformer with listener-dependent modeling, showcasing state-of-the-art performance in zero-shot evaluation of singing voice conversion samples. Nevertheless, as highlighted in the VoiceMOS Challenge 2023 \cite{10389763}, the objective to create a model capable of reliably predicting MOS in diverse real-world scenarios remains unfulfilled, making general MOS prediction in practical applications an ongoing area of research.


From our perspective, a significant limitation hindering the MOS prediction systems from applying to the real and open world is they almost \textit{exclusively focus on point estimates--neglect to provide uncertainty information}. Uncertainty estimation is an important building block of reliable, interpretable, and practical MOS prediction systems. On the one hand, uncertainty can help users to better understand the reliability of the forecast and make more rational decisions based on their needs. On the other hand, through the filtration of uncertainties, the model can reject unreliable predictions, thereby exhibiting stable performance. It is a crucial advantage given the current challenge of inconsistent MOS prediction accuracy and the need to address the diversity of samples in open-world scenarios.

The existing literature scarcely mentions uncertainty estimation in MOS prediction tasks, with only a few studies of weak relevance. Liang et al. \cite{liang23d_interspeech} propose a deep neural network-based approach to estimate the posterior distribution of the MOS, replacing the traditional point estimate method. But the mean of the posterior distribution could be considered a new form of point estimation and additional information within the distribution is overlooked in testing. Ravuri et al. \cite{Ravuri2023UncertaintyAA} demonstrate that the uncertainty obtained from SSL models exhibits a correlation with the MOS, facilitating the assessment of speech quality in low-resource settings. This uncertainty serves as a strategy to derive quality estimates, while the level of confidence associated with these estimates has yet to be determined. 

Given the significant yet underexplored impact of uncertainty on MOS prediction, we undertake a systematic discussion and investigation into the uncertainties inherent in MOS prediction tasks. Specifically, we analyze the sources of uncertainty in the MOS prediction, categorizing them into aleatoric and epistemic uncertainties. Building on this foundation, we propose an uncertainty-aware MOS prediction framework. This framework not only assesses speech quality but also quantifies the associated uncertainty. Heteroscedastic regression and Monte Carlo (MC) dropout techniques are employed for aleatoric and epistemic uncertainty, respectively. To the best of our knowledge, this is the first MOS prediction system that can make uncertainty estimation. We validate the effectiveness of our uncertainty modeling methods through experiments and explore the benefits of incorporating uncertainty into the application of MOS prediction models in the real and open world.

\section{Uncertainty in MOS Prediction}

\subsection{Category of Uncertainty}
Uncertainty is commonly classified into two types: aleatoric uncertainty and epistemic uncertainty \cite{Hllermeier2019AleatoricAE, hu2023uncertainty}. 
\begin{itemize}
\item Aleatoric uncertainty, or data uncertainty, refers to the inherent randomness or noise present within the data itself. It is irreducible and intrinsic to the data being analyzed. 

\item Epistemic uncertainty, on the other hand, accounts for the uncertainty in the model parameters. Generally, it is caused by a lack of knowledge of the neural network \cite{Gawlikowski2021ASO}. Unlike aleatoric uncertainty, epistemic uncertainty can be reduced when more information or data becomes available.
\end{itemize}

\subsection{Sources of Uncertainty}
The main sources of uncertainty in MOS prediction include:

\begin{enumerate}

\item \textbf{The diversity of synthetic speech.} Driven by the influence of synthesized content, the target speaker, and the synthesis algorithm, the produced speech exhibits a high degree of complexity and variability.

\item \textbf{The subjectivity of annotation.} MOS reflects people's subjective opinions on speech quality. Influenced by cultural background,  perceptual capabilities, and personal preference, the opinion annotations are highly uncertain. 

\item \textbf{Small-scale training data.} Gathering manually scored data is challenging due to limited datasets. This scarcity restricts models' ability to learn diverse speech variations, leading to reduced effectiveness and increased uncertainty.

\item \textbf{OOD test samples.} 
Due to the lack of external knowledge, models cannot interpret OOD samples and often produce unreliable and uncertain predictions.
\end{enumerate}

Of the four aspects discussed, the initial two are categorized under aleatoric uncertainty -- rooted in randomness, and the subsequent two under epistemic uncertainty -- stemming from knowledge limitations.  

\section{Methodology}

\begin{figure}[t]
  \centering
  \includegraphics[width=\linewidth]{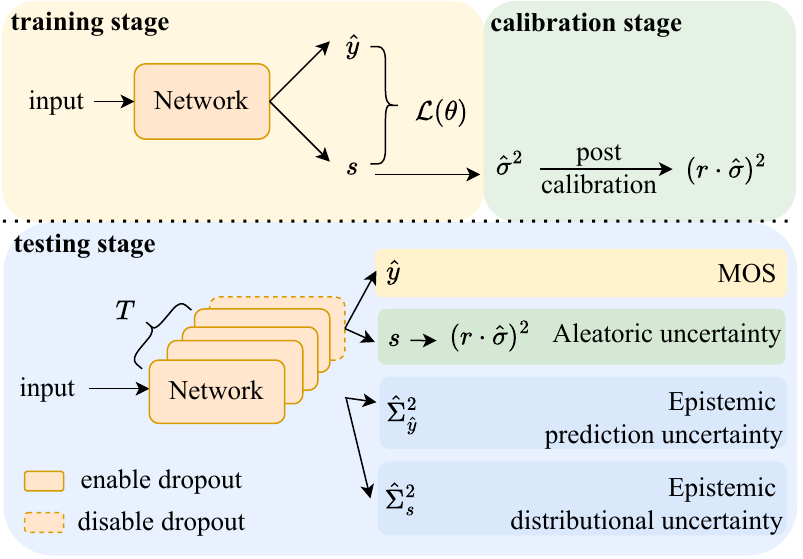}
  \caption{The pipeline of our uncertainty-aware MOS prediction system, including training, calibration, and testing stages.}
  \label{fig:pipeline}
\end{figure}

As shown in Figure~\ref{fig:pipeline}, the model simultaneously learns to predict speech quality MOS and capture aleatoric uncertainty in the training stage by heteroscedastic regression. The aleatoric uncertainty is further calibrated by scalar $r$ in the calibration stage. In the test stage, two kinds of epistemic uncertainty are modeled by the MC dropout, performing $T$ times forward propagation on the network with dropout enabled and collecting the output results to estimate uncertainty.

 \subsection{Aleatoric Uncertainty Modeling}

Aleatoric uncertainty can be further divided into homoscedastic uncertainty and heteroscedastic uncertainty \cite{NIPS2017_2650d608}. Homoscedastic uncertainty implies a consistent level of uncertainty across different inputs, heteroscedastic uncertainty refers to the uncertainty changes depending on different inputs. Data of MOS prediction tasks exhibits heteroscedasticity because the level of uncertainty differs across inputs which is influenced by the TTS system and the synthesized text. Therefore we model the aleatoric uncertainty as a data-dependent parameter $\sigma^2$. 

\textbf{Pre-estimation.} Following Kendal et al. \cite{NIPS2017_2650d608} and Laves et al. \cite{pmlr-v121-laves20a}, we also include $\sigma^2$ as one of the training targets of the model. Formally, the aleatoric uncertainty-aware model $f_\theta$ takes the data $x$ as input and outputs the target that follows the Gaussian distribution with mean equal to $\hat{y}(x)$ and variance $\hat{\sigma}^2(x)$, which is
%
\begin{align}
f_\theta(x)\sim\mathcal{N}\left(\hat{y}(x),\hat{\sigma}^2(x)\right).
\end{align}

We use the Negative Gaussian Log-Likelihood loss (NLL) as the optimization objective of the model. Given a training set $\mathcal{D}$, it can be expressed as:
\begin{align}
    \mathcal{L}(\theta)=&-\sum_{i=1}^D\log p(y_i|\hat{y}_i,\hat{\sigma}_i^2) \\
    =&\sum_{i=1}^{D}(\frac{\log(\hat{\sigma}_i^2)}2+\frac{(y_i-\hat{y}_i)^2}{2\hat{\sigma}_i^2}+\mathrm{const}),
    \label{eq:nll}
\end{align}
where $D=|\mathcal{D}|$ and $y_i$ denotes the label. In practice, we directly estimate $s_i:=log(\hat{\sigma}_i^2)$ to avoid numerical instability. 

\textbf{Post-calibration.} Based on obtaining a preliminary estimate of aleatoric uncertainty through heteroscedastic regression, we further calibrate the uncertainty by introducing a simple scalar parameter $r$. More specifically, after training the model $f_\theta$, we adjust its uncertainty on the calibration set $\mathcal{C}$. The adjusted uncertainty is denoted as $(r\cdot\hat{\sigma})^2$, which is derived by applying a scaling factor $r$ to the original uncertainty $\hat{\sigma}^2$. Thus the probability density function of the target value is 
$\mathcal{N}(\hat{y}(x),(r\cdot\hat{\sigma}(x))^2)$.

Substituting the new standard deviation $r\cdot\hat{\sigma}(x)$ into the NLL loss function, we obtain the optimization objective for finding the best scaling factor $r$ :
\begin{align}
\mathcal{L}(r)=&-\sum_{i=1}^C\log p(y_i|\hat{y}_i,(r\cdot\hat{\sigma}_i)^2)  \\
=&C\log(r)+\frac12r^{-2}\sum_{i=1}^C\frac{(y_i-\hat{y}_i)^2}{(\hat{\sigma}_i)^{2}}+\mathrm{const},
\end{align}
where $C=|\mathcal{C}|$. Since the model $f_\theta$ is fixed during the calibration stage, $\hat{y}$ and $\hat{\sigma}$ are treated as known constants. By setting the derivative of $\mathcal{L}(r)$ with respect to $s$ to zero, we find the optimal value of $r$ is:
\begin{align}\label{eq:eqr}
r = \pm\sqrt{\frac{1}{C}\sum_{i=1}^{C}\frac{(y_i-\hat{y}_i)^2}{(\hat{\sigma}_i)^2}},
\end{align}
which reflects how the scaling factor minimizes the entire loss function under the given dataset and model prediction uncertainty. Therefore, by Equation~\ref{eq:eqr}, we can simply obtain the calibration coefficients without any iterative optimization process. 
\subsection{Epistemic Uncertainty Modeling}
\label{sec:Epistemic Uncertainty}

To capture the epistemic uncertainty $\Sigma^2$, we use MC dropout \cite{pmlr-v48-gal16}. Dropout is a popular regularization technique that prevents over-fitting by randomly disabling a portion of neurons during the training time. Gal and Hahramani \cite{pmlr-v48-gal16} interpret dropout training in networks as approximate Bayesian inference in deep Gaussian processes and propose to enable dropout at test time to quantify the uncertainty.

In our system, the model is trained with dropout. At the test time, we run $T$ stochastic forward passes with dropout enabled at a given probability $p$. Test time outputs can be seen as samples from the approximate Bayesian. Thus we use the variance of the results as the predictive epistemic uncertainty. Corresponding to the two outputs of the model: the speech quality $y$ and the log of variance $s$, the epistemic uncertainty is modeled from two perspectives: the epistemic prediction uncertainty $\hat{\Sigma}_{\hat{y}}^2$ and the epistemic distributional uncertainty $\hat{\Sigma}_s^2$.

\textbf{Epistemic prediction uncertainty.} The epistemic uncertainty associated with the prediction, $\hat{y}$, quantifies the variability in the model's quality prediction outcomes. It measures the reliability of the model in quality prediction for a given input. This concept is formalized as: 
\begin{align}
\hat{\Sigma}_{\hat{y}}^2=\frac1T\sum_{t=1}^T\left(\hat{y}_t-\frac1T\sum_{t=1}^T\hat{y}_t\right)^2.
\end{align}

\textbf{Epistemic distributional uncertainty.} Since $s$ is actually a feature related to data, this type of uncertainty quantifies the model's confidence in capturing the underlying distribution of the input data. A high level of epistemic distributional uncertainty signifies the model's inadequate comprehension of the data. The quantification of epistemic uncertainty for $s$ is:
\begin{align}
\hat{\Sigma}_s^2=\frac1T\sum_{t=1}^T\left(s_t-\frac1T\sum_{t=1}^Ts_t\right)^2.
\end{align}
%

\begin{figure}[t]
  \centering
  \includegraphics[width=\linewidth]{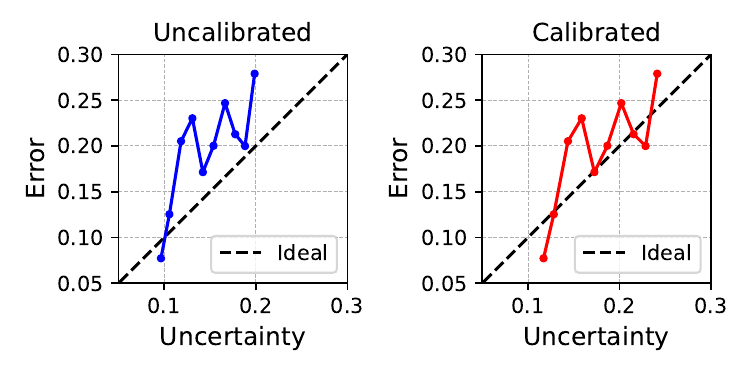}
  \vspace{-12pt}
  \caption{The error-uncertainty curve before and after calibration in BVCC test set.}
  \vspace{-12pt}
  \label{fig:cal}
  
\end{figure}

\section{Experiments}

\subsection{Experiments Setup}

\textbf{Datasets.} In our experiments, we utilize the BVCC \cite{cooper21_ssw} for training, which comprises 7,106 English audio samples from 187 TTS and Voice Conversion (VC) systems included in previous challenges, as well as samples from the ESPnet-TTS \cite{Hayashi2020EspnetTTSUR}. For OOD testing, we employ VCC2018 \cite{bee295ff344749e98ed084b97145216d}, which consists of English VC system outputs from Voice Conversion Challenge 2018 and has minor overlap with BVCC. Additionally, we use BC2019 \cite{Wu2019TheBC}, which contains Mandarin TTS samples from the 2019 Blizzard Challenge, and TMHINT-QI-II \cite{zezario2023study}, released for the VoiceMOS challenge 2023, comprising noisy and enhanced speech from speech enhancement systems.

\textbf{Architecture.} We use wav2vec 2.0 base model \cite{baevski2020wav2vec} as the backbone. After feature extraction, we compress the temporal dimension by averaging and then apply a linear layer to reduce it from 768 to 256. Subsequently, we attach two task-specific heads: one for quality scores $\hat{y}$ and another for $s$. Each head includes a dropout layer followed by two linear layers.

\textbf{Implementation details.} The models are trained with a batch size of 8 and a learning rate of 0.0003. The calibration phase is conducted on the validation set. In estimating epistemic uncertainty using MC dropout during testing, a dropout rate $p$ of 0.5 and $T=25$ executions are employed.

\subsection{Evaluation Metrics}

We evaluate the performance from two aspects: quality prediction and uncertainty estimation. The quality prediction aspect utilizes Mean Squared Error (MSE) and system-level Spearman's Rank Correlation Coefficient (SRCC). For uncertainty estimation, we employ the following metrics:

\textbf{Negative Log-likelihood (NLL).} The NLL measures the fit of the model to the data. Its calculation follows the Equation~\ref{eq:nll}. 

\textbf{Uncertainty calibration error (UCE).} The metric assesses the data by dividing it into M equal-width bins ${B_1, B_2, ..., B_m}$ based on uncertainty levels, then calculates the average of the differences between each bin's error and its uncertainty \cite{pmlr-v121-laves20a} :
\begin{align}
    \mathrm{UCE}:=\sum_{m=1}^M\frac{|B_m|}n|\mathrm{err}(B_m)-\mathrm{uncert}(B_m)\big|,
\end{align}
where $n$ is the number of inputs and $M$ is 10 in experiments. 

%

\textbf{Sharpness.} It measures a model's overall uncertainty level on a dataset through the average of uncertainties of all samples in dataset \cite{pmlr-v80-kuleshov18a}.

\textbf{Area Under Curve (AUC).} It's used to assess the performance of in-domain and OOD binary classification.

\subsection{Experimenting with Aleatoric Uncertainty Modeling}

\begin{figure}[t]
  \centering
  \includegraphics[width=\linewidth]{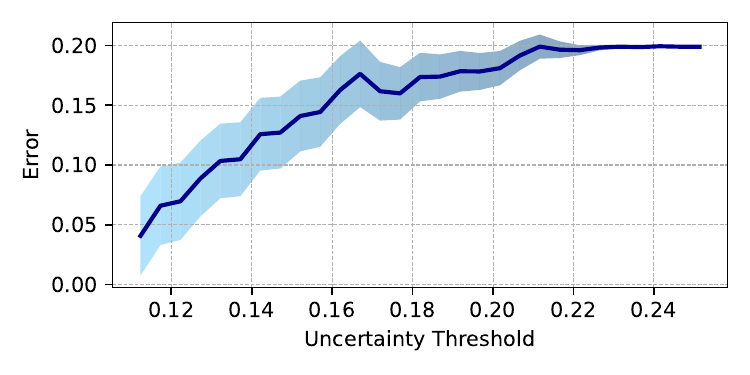}
  \caption{The relationship between uncertainty threshold and MSE of the reliable subset in the BVCC test set.}
  \label{fig:scales}
\end{figure}

We first verify the effectiveness of heteroscedastic regression, denoted as HR, by making a comparison with MSE loss and another aleatoric uncertainty modeling method, denoted as KL \cite{zerva-etal-2022-disentangling}. MSE is commonly used in MOS prediction, yet lacks the capability for uncertainty estimation. The KL method models annotator scores as a target Gaussian distribution, minimizing KL divergence to reduce the distance between target and predicted distributions. Table~\ref{tab:e1} displays the average results of each method across eight seeds. Our method not only surpasses the KL model across all metrics but also achieves competitive results in quality prediction when compared with the MSE model. This demonstrates the HR model's strong quality prediction ability and capability to model aleatoric uncertainty effectively.
 
\begin{table}[th]
\caption{Results of aleatoric uncertainty modeling methods.}
\label{tab:e1}
\centering
\begin{tabular}{ccccc}
\toprule
\textbf{Model} & \textbf{UCE $\downarrow$} & \textbf{NLL $\downarrow$} & \textbf{MSE $\downarrow$}& \textbf{SRCC $\uparrow$} \\ 
\midrule
MSE & - & - & 0.223 & 0.930  \\
KL & 0.7523 & 1.049 & 0.254 & 0.927  \\
HR & 0.0456 & 0.657 & \textbf{0.203}& \textbf{0.932} \\
\midrule
calibrated HR& \textbf{0.0338} & \textbf{0.632} & \textbf{0.203}& \textbf{0.932} \\
\bottomrule
\end{tabular}
\end{table}

We further validate the effectiveness of the calibration stage. The last two rows in Table~\ref{tab:e1} show that simple scaling calibration significantly enhances the UCE and NLL metrics, with $p=0.0168$ and $p=0.0372$, respectively. The fact that the MSE and SRCC metrics remain unchanged is attributed to the calibration solely addressing uncertainty and not modifying the quality prediction. In Figure~\ref{fig:cal}, we show the error-uncertainty curves before and after calibration. The ``ideal'' curve represents perfect uncertainty estimation, that is, the uncertainty accurately reflects the model's prediction error. The ``calibrated'' curve is closer to the ideal one than the ``uncalibrated'' curve, demonstrating the effectiveness of our calibration method.

Building on our advancements in accurately modeling aleatoric uncertainties, we apply the forecasted aleatoric uncertainty for selective prediction. Specifically, we determine an uncertainty threshold, discarding predictions beyond this value as unreliable, and subsequently assess the model’s performance on the deemed reliable subset, as illustrated in Figure~\ref{fig:scales}. The shaded area in the figure shows the proportion of unreliable samples, and the curve points indicate the MSE of the reliable subset at various thresholds. As the threshold increases, so does the MSE, demonstrating that aleatoric uncertainty effectively captures accuracy and supports the viability of using uncertainty to manage model performance in practice.

\begin{figure}[t]
  \centering
  \includegraphics[width=\linewidth]{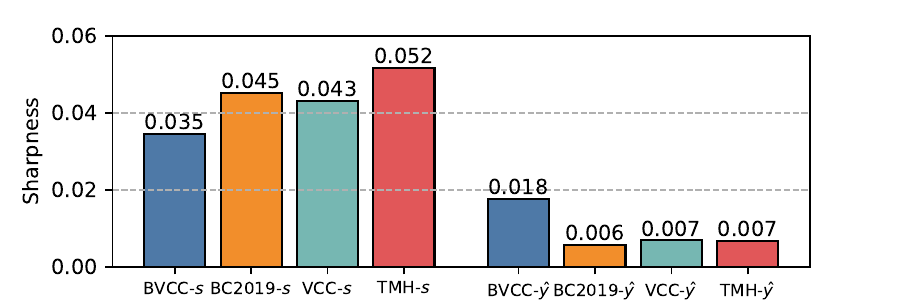}
  \caption{The sharpness for epistemic distributional uncertainty and epistemic prediction uncertainty across for test sets: BVCC, BC2019 (OOD), VCC2018 (OOD), and TMHINT-QI-II (OOD).}
  \vspace{-12pt}
  \label{fig:ood1}
\end{figure}

\begin{figure}[t]
  \centering
  \includegraphics[width=\linewidth]{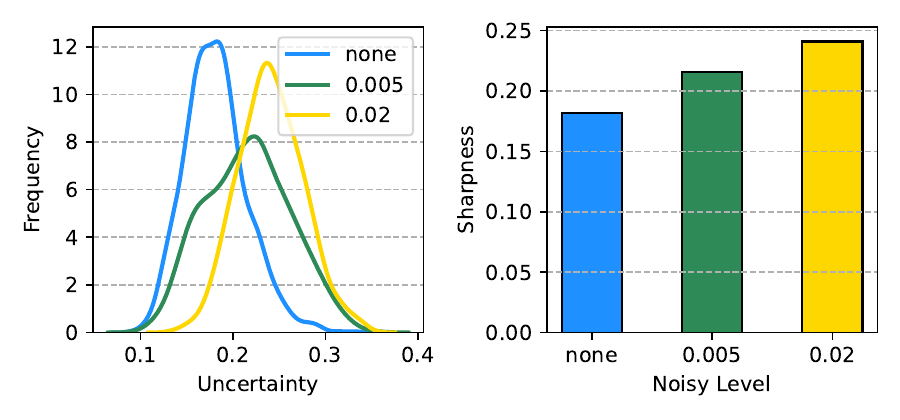}
  \caption{The distribution and sharpness of epistemic distributional uncertainty on data with different noise levels.}
  \vspace{-12pt}
    \label{fig:ood2}
\end{figure}

\subsection{Experimenting with Epistemic Uncertainty Modeling}

We set two OOD scenarios to evaluate the effectiveness of the predictive epistemic uncertainty. We first test the BVCC with three OOD datasets and compare two kinds of epistemic uncertainty captured in Figure~\ref{fig:ood1}. These two types of epistemic uncertainty exhibit inconsistent trends and magnitudes: the sharpness of epistemic distributional uncertainty is larger overall and more sensitive to OOD samples, while the epistemic prediction uncertainty is the opposite. We analyze that this is because of the difference caused by the difficulty and target value range of the two output heads to learn. Consequently, epistemic distributional uncertainty proves more effective for OOD detection. We further investigate using this uncertainty for OOD classification, with results in Table~\ref{tab:ood}. It shows strong performance on all three OOD datasets, particularly TMHINT-QI-II, likely due to its significant difference from the training domain.

\begin{table}[th]
\caption{AUC of OOD detection, which refers to using uncertainty as a basis for classifying the mix BVCC and OOD data.}
\label{tab:ood}
    \centering
    \begin{tabular}{ccccccc}
    \toprule
  &  \multicolumn{2}{c}{\textbf{BC2019}}&  \multicolumn{2}{c}{\textbf{VCC2018}}&  \multicolumn{2}{c}{\textbf{TMHINT-QI-II}}\\ 
    \midrule
    \textbf{AUC $\uparrow$} & \multicolumn{2}{c}{0.641}  & \multicolumn{2}{c}{0.646} & \multicolumn{2}{c}{0.768}\\
    \midrule\midrule
    &  \multicolumn{3}{c}{\textbf{noise level of 0.005}}&  \multicolumn{3}{c}{\textbf{noise level of 0.02}}\\
     \midrule
    \textbf{AUC $\uparrow$} & \multicolumn{3}{c}{0.723}&  \multicolumn{3}{c}{0.890}\\
    \bottomrule
\end{tabular}

\end{table}

The second OOD evaluation scenario examines BVCC test set with different noise levels: none, 0.002, and 0.01, shown in Figure~\ref{fig:ood2}. The results indicate the sharpness of the model's epistemic distributional uncertainty grows with noise increase, highlighting its sensitivity to OOD features. The lower part in Table~\ref{tab:ood} shows that using this uncertainty for OOD classification tasks offers promising performance, significant for applying the MOS prediction system in real and open-world settings.

\subsection{Discussion}
In these experiments, we conduct a preliminary analysis that showcases the potential applications of aleatoric and epistemic uncertainties. Accurately capturing aleatoric uncertainty in MOS prediction enables model performance understandable, and controllable by setting of appropriate thresholds to do selective prediction. Meanwhile, epistemic uncertainty displays a heightened sensitivity to OOD data, providing a viable approach for OOD detection. Equipped with selective prediction and OOD detection capabilities, the MOS prediction system can be reliably applied in real and open-world environments.

\section{Conclusion}

In this paper, we emphasize the importance of uncertainty modeling for a practical MOS prediction system and analyze the uncertainty sources of MOS prediction tasks. We then model aleatoric uncertainty and epistemic uncertainty using calibrated heteroscedastic regression and MC dropout, respectively. The experimental results show that our uncertainty-aware MOS system not only performs well but also holds promise for applications in selective prediction and OOD detection.

\section{Acknowledgements}
This work has been supported by the National Key R\&D Program of China through grant 2022ZD0116307 and NSF China (Grant No.62271270).

\bibliographystyle{IEEEtran}
\bibliography{mybib}

\end{document}